\documentclass{article}

\usepackage{arxiv}

\usepackage[utf8]{inputenc} 
\usepackage[T1]{fontenc}    
\usepackage{hyperref}       
\usepackage{url}            
\usepackage{booktabs}       
\usepackage{amsfonts}       
\usepackage{nicefrac}       
\usepackage{microtype}      
\usepackage{lipsum}		
\usepackage{graphicx}
\usepackage{natbib}
\usepackage{doi}
\usepackage{booktabs} 
\usepackage{multirow} 

\title{A text autoencoder from transformer for fast encoding language representation}

\author{
    Tan Huang\\
	Nanyang Technological University\\
	Singapore 639798 \\
	\texttt{thuang006@e.ntu.edu.sg} \\
}

\date{}

\hypersetup{
pdftitle={A template for the arxiv style},
pdfsubject={q-bio.NC, q-bio.QM},
pdfauthor={David S.~Hippocampus, Elias D.~Striatum},
pdfkeywords={First keyword, Second keyword, More},
}

\begin{document}
\maketitle

\begin{abstract}
In recent years BERT shows apparent advantages and great potential in natural language processing tasks. However, both training and applying BERT requires intensive time and resources for computing contextual language representations, which hinders its universality and applicability. To overcome this bottleneck, we propose a deep bidirectional language model by using window masking mechanism at attention layer. This work computes contextual language representations without random masking as does in BERT and maintains the deep bidirectional architecture like BERT. To compute the same sentence representation, our method shows O(n) complexity less compared to other transformer-based models with O($n^2$). To further demonstrate its superiority, computing context language representations on CPU environments is conducted, by using the embeddings from the proposed method, logistic regression shows much higher accuracy in terms of SMS classification. Moverover, the proposed method also achieves significant higher performance in semantic similarity tasks.

\end{abstract}


\section{Introduction}
The basic task of NLP is to build a language model, based on which most of the NLP applications such as machine translation can be carried out (ref). Recently, BERT (Bidirectional Encoder Representations from Transformers) and its variations have brought significant improvements in learning natural language representation, and they have achieved state-of-the-art performances on various downstream tasks such as GLUE benchmark and question answering. This success of BERT continues in various unsupervised tasks such as the N-best list reranking for image captioning and NMT, demonstrating the superiority of bidirectional language models for unsupervised tasks.

However, when training the BERT model, it usually requires huge corpus and large parameters, which lead to usage of expensive hardware equipment and long duration of training time. During training, the BERT uses the masked language modeling (MLM) objective, which is to predict the original ids of explicitly masked words from the input. Moreover, due to the random masking mechanism of BERT, for each input sentence, 70\% of the tokens in the sentence are randomly masked, only 30\% of tokens in the sentence can be trained in one epoch of training. This makes the BERT model training is extremely inefficient. Therefore, it is necessary to increase the training efficiency of BERT model where domain specific applications requires a fast set up and validation. 

In this paper, we try to solve the above limitation by proposing a novel bidirectional language model named window masking model. The proposed model is trained with a new learning objective named next token prediction based on the token preceding it. This method uses the text preceding the target word as input text to predict the target word, namely, the concatenation of embedding from attention layer where target word is masked and embedding from previous residual input will be used to predict the target word. To learn the proposed objective, we devise a window masking operation and an next prediction mechanism in the residual layer inside the model based on the Transformer encoder (Vaswani et al.,2017). These components enable the proposed model to compute contextualized language representations at once while maintaining the advantages of the deep bidirectional architecture of language model.

We conduct a series of experiments on two unsupervised tasks: the N-best list reranking and the unsupervised semantic textual similarity. First, in the training process with GPU, we show that the proposed method’s loss is 5 times smaller than the BERT-based model given the same training epochs. Second, even with this faster training process, the proposed method achieves competitive performances to BERT on reranking tasks and unsupervised semantic textual similarity tasks.

\section{Related Works}
\label{sec:headings}

Rumelhart et al\citep{rumelhart1986learning} proposed that hidden units of neural network can represent features for a specified task after weight adjustment by using back-propagation. This methodology has been well adapted for use in natural language processing domain for developing language models for word embedding. Prior to distributed word representations based on neural network, statistical language model is trained by computing the joint probability function of word sequences, which usually cause curse of dimensionality during testing because of unseen words in training set. Distributed representation is proposed to overcome this limitation like this, for word sequence containing word not seen in training set but similar to words in a seen sentence, the target word sequence can still obtain high probability because of similar word vectors with that of sentence in the training set\citep{bengio2003neural}. Moreover, Mikolov et al\citep{mikolov2013distributed} produced many works to distributed representation such as Skip-gram, negative sampling, all of which make word embedding in NLP an essential methodology and corner stone in many tasks including machine translation speech recognition.

However, the above-mentioned word level representations cannot capture information arising from various polysemy of word use across linguistic context. To address this issue, Peters et al\citep{peters2018deep} trained bidirectional LSTM with a coupled language model objective for producing word vectors, which led to that the derived representations improved the state of the art in diverse language understanding problems. Radford et al\citep{radford2018improving} proposed a framework with transformer as base architecture for achieving long-range dependency, the ablation study shows that apparent score drop without using transformers. All of the results show that contextualized representation are beneficial in language modelling. Furthermore, Devlin et al\citep{devlin2018bert} devised bidirectional encoder representation transformers which show significant progress in eleven natural language processing tasks.

Although deep contextualized language model achieved excellent performance, huge amount of parameters of these models incur extremely high cost of computing hardware and computation time. Another line of research tries to address this problem. Lan et al\citep{lan2019albert} proposed A Lite BERT architecture which has 18 times fewer parameters and 1.7 times faster tranning time than a traditional BERT architecture. While Sanh et al\citep{sanh2019distilbert} presented a method to pretrain a smaller model that can be finetuned for the downstream task, and achieved a 1.4 times fewer parameter with 1.6 times faster inference.

However, none of these studies tried to investigate the effect of attention layer’s information leakage, for the purpose of information recombination, on the language model’s training speed and language representation efficacy for the purpose of decreasing computational resources utilization and training time.

\section{Language Model Benchmark}
\label{sec:others}
The training of traditional language model is a process where the targeted token (i-th token) is predicted based on its preceding text as input context. In this work, we use the model of seq2seq with beam search decoder to produce the original list for reranking. We use GPT, as a unidirectional language model, is used as the baseline model for language performance comparison. In the process of training the GPT, the targeted token is predicted only based on the preceding tokens’ embedding in the attention layer, which means the given token does not encapsulate the information from its following tokens.

In 2018, bidirectional encoding representation from transformers (BERT) is proposed. BERT tries to obtain bidirectional language representation by adopting a masked language model mechanism. For each sentence feeding into the model, some tokens in the input sentence will be selected randomly and masked (the randomly masked ratio is fixed), and the objective is to predict the original token based on its surrounding tokens. Therefore, BERT is a bidirectional language model. Apparently, a given sentence should be fed into the model for several times in order to make each token well trained, because effective training only occurs at the masked position. Therefore, the conventional way of training bidirectional language model is both time-consuming and expensive in terms of hardware.

\section{Proposed Method}

\subsection{Window masking based attention}
To learn the bidirectional representation from language models, we develop a window masking mechanism in the attention layer for prediction of each token based on its preceding tokens in the residual connection layer.

\begin{figure}[htp]
    \centering
    \includegraphics[width=8cm]{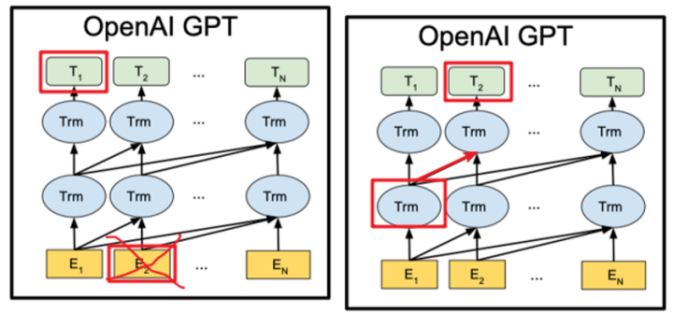}
    \caption{The masking mechanism of GPT}
    \label{fig:galaxy1}
\end{figure}

\begin{figure}[htp]
    \centering
    \includegraphics[width=8cm]{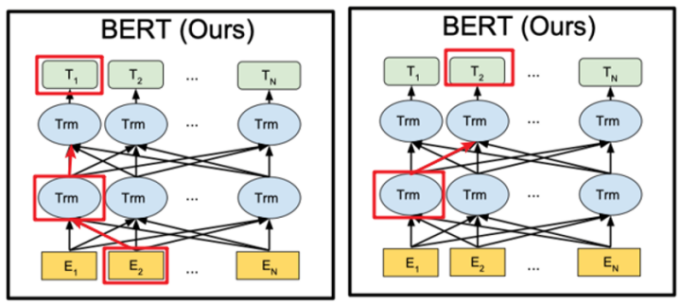}
    \caption{The masking mechanism of BERT: any token is randomly masked with a probability of 30\%, the masked token is replaced by the specified place holder like [mask]}
    \label{fig:galaxy2}
\end{figure}

\begin{figure}[htp]
    \centering
    \includegraphics[width=8cm]{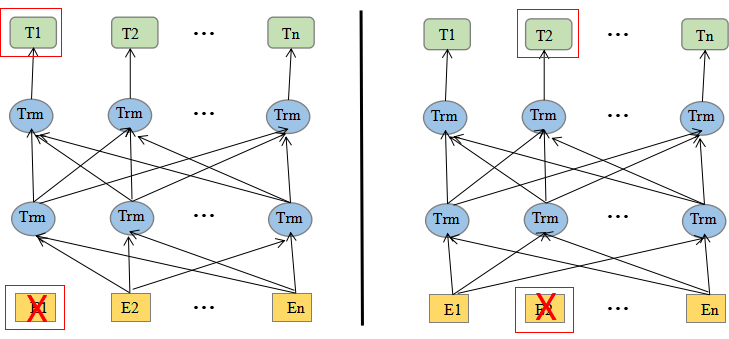}
    \caption{The window masking: when predicting T1, E1 is masked (left); when predicting T2, E2 is masked (right)}
    \label{fig:galaxy3}
\end{figure}

\subsection{Causal prediction in residual connection layer}
The transformer in language models usually includes attention layers followed by residual connection layer, the residual connection layer is used for efficient training of deep neural networks. In this work, targeted token is predicted based on its preceding tokens in the residual connection layer, as shown in Figure 4. In this way, the model to be trained will not know the information of the targeted token when predicting the targeted token, the purpose of doing so is to avoid the residual connection information leakage.

\begin{figure}[htp]
    \centering
    \includegraphics[width=8cm]{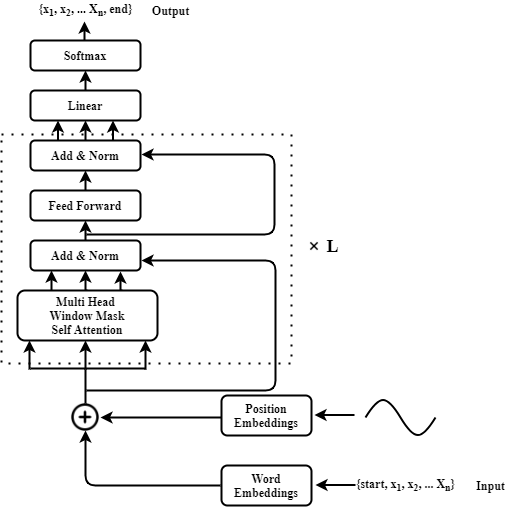}
    \caption{Architecture of our TDT}
    \label{fig:galaxy4}
\end{figure}

\subsection{Window masking mechanism}
As shown in Figure 4, in all attention layers the values of the selected and targeted token to be predicted is masked, this window masking operation is to prevent the information of the targeted token going to next attention layer. However, when this operation is used in multi-layer transformers, from first attention layer to second attention layer, scaled dot-product operation between different tokens (Q and K) will let some of the information of targeted token flowing to the output layer, how much of the information is flowing through depends on the relation between the tokens. And we have studied and compared the effect of various information flowing on the language’s performance,the results show that window masking mechanism is beneficial for language model to learn the fully contextualized representations, while keeping fast training speed. Notably, in this work, our aim is to put forward an efficient method for training language models which can learn tokens’ essential features and compute bidirectional language representations in a word level in an unsupervised way, while ignoring signal noise. Therefore, the information flowing of the targeted token after first attention layer is useful for efficient data coding, which is good for downstream unsupervised application of the as trained language models.

\begin{figure}[htp]
    \centering
    \includegraphics[width=8cm]{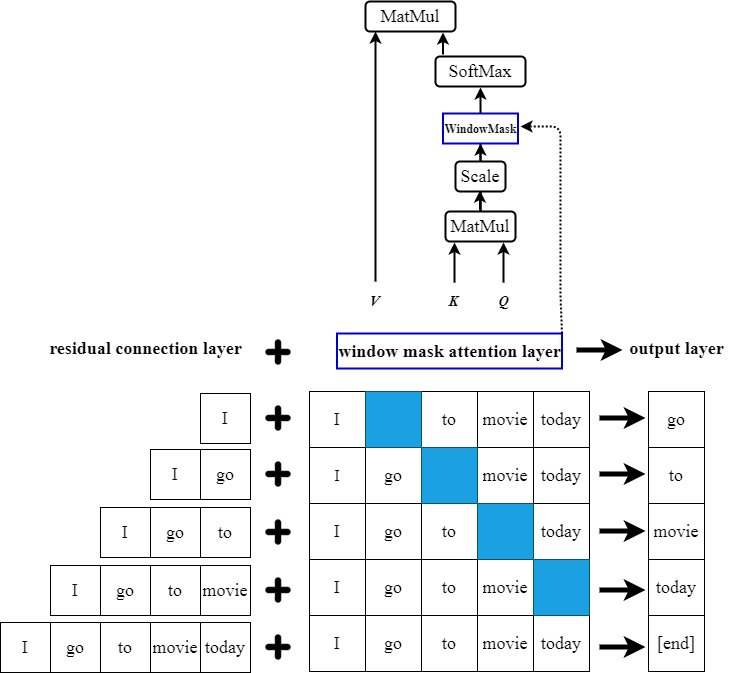}
    \caption{Window masking of the scaled dot-product attention mechanism by using the example of  the sequence: “I go to movie today”}
    \label{fig:galaxy5}
\end{figure}

\section{Experiments}

To evaluate the proposed method, we conduct a series of experiments. Firstly, we evaluate the contextual language representations obtained from the window masking model on the N-best list reranking tasks. We subsequently apply our method to unsupervised semantic textual similarity (STS) tasks. The following sections will demonstrate that the proposed model is much faster to reach convergence than the BERT does during training while showing comparative or even better accuracy than those of the BERT on reranking tasks and STS tasks.

\subsection{Language Model Setups}
This work uses the GPT and conventional BERT as the benchmark models. GPT is unidirectional language model, and BERT is bidirectional language model. Therefore, we compare TDT with GPT to show the advantages of language models arising from bidirectional learning; we compare TDT with BERT to show the training efficiency of the newly proposed method. 

For a fair comparison, each model has the same number of parameters based on the Transformer as followed: number of attention layers (L) is 3, embedding dimensions for input sequence (dim) is 64, number of attention head is 4, the number of hidden units is 2048 for the position-wise feed-forward layers. Relu activation is used in the feed forward layers. Position embeddings have been trained during training process with 64 tokens as the maximum sequence length. The vocabulary size (V) is about 160 000.

For training, we make a training instance consisting of a single sentence with [start] and [end] before the first token and after the last token, respectively. The size of training batch is 64 sentences per batch, and train language models 5000 steps for machine translation task. The language model is trained with Adam optimizer with an initial learning rate of 0.0001, the dropout rate on all layers is 0.2.

To train language models that we implement, we use about 16 GB latest English Wikipedia dump that has about 130M sentences. The trained models are used for reranking in neural machine translation (NMT) and unsupervised semantic textual similarity tasks.

\subsection{Training and Inference Time Analysis}
For a specified language model, both its training time and running time at inference stage accounts significantly for its large scale deployment in real world applications. In this work, we compare the running time of obtaining language representations of a given sentence for each language model. In terms of unsupervised semantic textual similarity task, output from embedding layer, attention layer and last fully connected layer are all used separately to represent the corresponding sentences. While for reranking task, the output from the last layer is fed into softmax function in order to compute the likelihood of each token.

\begin{figure}[htp]
    \centering
    \includegraphics[width=8cm]{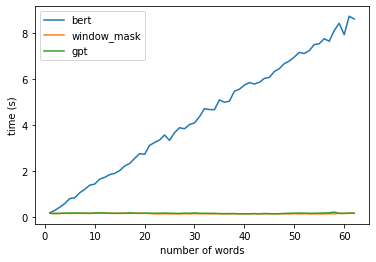}
    \caption{Average runtimes of window-masking model according to the number of words on reranking tasks}
    \label{fig:galaxy6}
\end{figure}

\begin{figure}[htp]
    \centering
    \includegraphics[width=8cm]{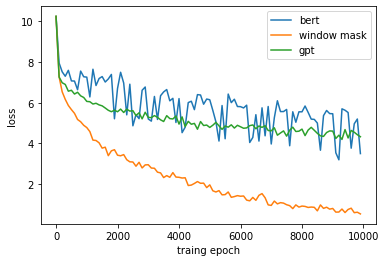}
    \caption{Loss evolution during training process}
    \label{fig:galaxy7}
\end{figure}

To measure the runtime, the processor specification is: Intel(R) Core(TM) i5-5200U CPU @ 2.20GHz, the CPU is with 2 cores, 4 logical processors and the TensorFlow 2.3.0 library with Python 3.8.3, the operating system is Ubuntu 18.04 LTS.To obtain the running time of each language model for computing the language representation given a text sequence with specified length, 20 parallel experiments are conducted and produce the averaged results. Figure 5 shows that the run-time of the TDT is faster than that of the conventional BERT, and it becomes significant as the sentence is longer. 

For fair comparison, the length of the sequence is fixed at 20 (usually the average length of sentence in modern English). Under this condition, in the reranking task, TDT cost 5.28 fold computing time less than BERT does, mainly because of the repetition of mask and prediction for each token.

\subsection{Word representations from language models for SMS classification}

The SMS Spam Collection v.1 is a public set of SMS labeled messages that have been collected for mobile phone spam research. It has one collection composed by 5,574 English, real and non-enconded messages, tagged according being legitimate (ham) or spam.

\begin{figure}[htp]
    \centering
    \includegraphics[width=8cm]{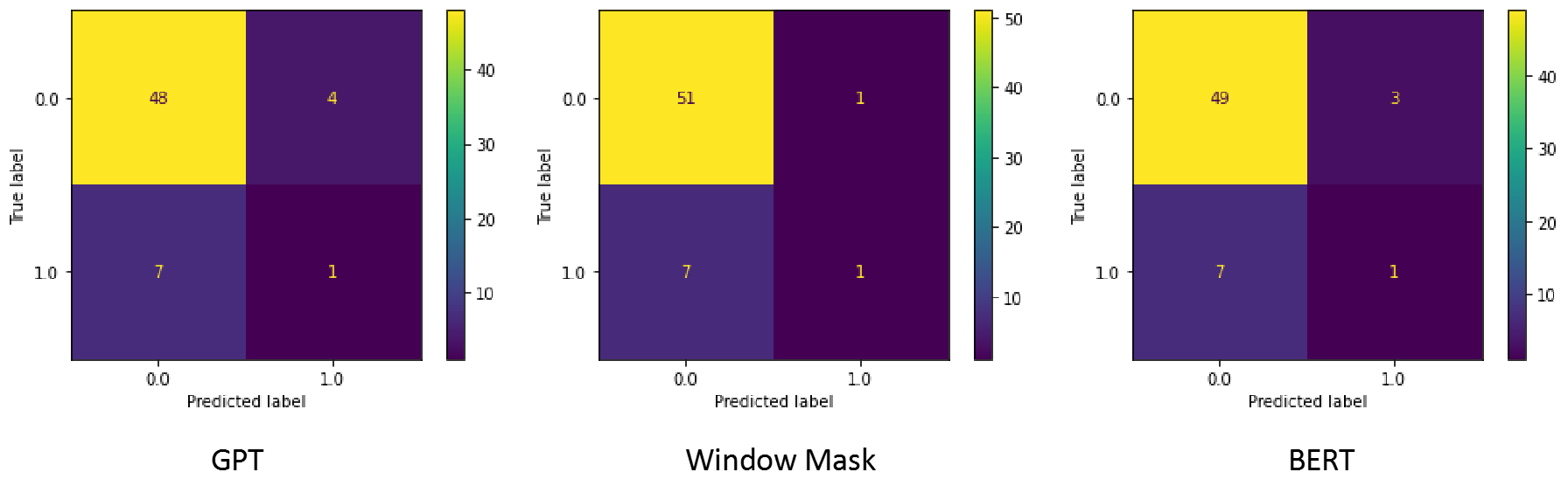}
    \caption{GPT, Window mask and BERT produce word representations for SMS classification, confusion matrix}
    \label{fig:galaxy8}
\end{figure}

\begin{table}[!htbp]
\centering
\caption{Prediction accuracy by using word representations for SMS classification}
\begin{tabular}{cc} 
\toprule
\multicolumn{1}{c}{Method}&{Accuracy}\\  

\hline
{Window mask}&0.87\\

\cline{1-2}  
{BERT}&0.83\\
\cline{1-2}
{GPT}&0.82\\

\bottomrule
\end{tabular}
\end{table}

\subsection{Word representations from language models for image captions reranking}

(to be tested)

\subsection{Reranking the N-best list}

To evaluate language models, we conduct experiments on the unsupervised task of reranking the N-best list. In the experiments, we apply each language model to rerank the 20-best candidate sentences, which are obtained in advance using each sequence-to-sequence model on NMT. The NMT models we implement is typical seq2seq with attention and beam search decoder.

The reranking score is calculated by linearly interpolating scores of beam search decoder and language models following the equation below:

                           Reranking score = {$\lambda$}*score\_{s2s} + {$\lambda$}*score\_{lm}

Scores2s is the score reported by beam search decoder for each candidate sentence, Scorelm is the the sum of log-likelihood of each token,  is the interpolation weight, which is fixed at 0.5.

\subsubsection{Reranking with each language model on Anki dataset}
Spanish to English translation task. The datasets can be accessed in  the website of Tab-delimited Bilingual Sentence Pairs (http://www.manythings.org/anki/).

It shows that BERT is much better than the others in reranking on NMT, it seems that the N-best hypotheses of the NMT model are more subtle to distinguish than those of other tasks from the language model perspective, because the base seq2seqNMT system for Spa→En is already quite good. All reranking results in the task of NMT demonstrate that the proposed TDT has similar performance with that of GPT and BERT.

\begin{table}[!htbp]
\centering
\caption{BLEU scores after reranking with each language model on Anki dataset}
\begin{tabular}{cc} 
\toprule
\multicolumn{1}{c}{Method}&{Spa-En}\\  
\hline
{seq2seq}&{0.529}\\
\hline
{Window mask}&0.526\\

\cline{1-2}  
{BERT}&0.528\\
\cline{1-2}
{GPT}&0.525\\

\bottomrule
\end{tabular}
\end{table}

\subsection{Unsupervised semantic textual similarity}

Following the reranking task, the test of semantic textual similarity (STS) is used to study each language model’s capability to learn contextual language representations. In this test, each language model calculate the semantic similarity of sentence pairs based on the language representation of each sentence extracted from it. We use STS Benchmark (ref) and SICK (ref), the two datasets have a set of sentence pairs with corresponding similarity scores. The evaluation metric of STS is the Pearson’s r between the predicted similarity scores by each language model and the labeled similarity scores of the given sentence pairs.

We conduct the task of unsupervised STS by comparing language models which are trained on the corresponding training datasets. For each language model to obtain the similarity score of a given sentence pair, outputs from attention layers of each model is denoted as contextual representation, the cosine similarity of the contextual language representations of the given sentence pair is used as the similarity score. For reference, outputs of the embedding layer (denoted as embed) and last fully connected layer (denoted as output) are also used to compute the similarity score. 

We used the GPT model trained on the same training datasets as a baseline for STS tasks. In our work, each language model is compared on their ability to extract token-level language representations.

\subsubsection{Results on STS benchmark}

The STS Benchmark datasets contains about 5749/1500/1379 pairs of similar and dissimilar sentences along with their semantic similarity scores (ranging from 0 to 5) in the corresponding train, dev and test datasets. The following table shows our method trained with the window masked attention learn the semantics of a sentence over the Transformer-based language models. The result shows that TDT has obvious better representation ability when using context vectors and output vectors than the other transformer based language models. Notably, in most cases, the the context representation show far better performances thatn embed and output representations, which demonstrate the directly data encoding efficacy of attention layers as an encoder.

\begin{table}[!htbp]
\centering
\caption{Pearson’s r results on STS benchmark dataset}
\begin{tabular}{ccccccc}
\hline

\multicolumn{1}{c}{ \multirow{2}*{Method} }& \multicolumn{3}{c}{STSb-test} &\multicolumn{3}{c}{STSb-dev}\\
\cline{2-7}
\multicolumn{1}{c}{}&{context}&{embed}&{output}&{context}&{embed}&{output}\\
\hline
\multicolumn{1}{c}{GPT}&0.075&0.12&0.023&0.12&0.072&0.023\\
\cline{1-7}
\multicolumn{1}{c}{Window mask}&0.23&0.14&0.066&0.28&0.079&0.15\\
\cline{1-7}
\multicolumn{1}{c}{Transparent BERT}&0.12&0.13&0.05&0.14&0.070&0.17\\

\hline

\end{tabular}
\end{table}

\subsubsection{Results on SICK(Sentences Involving Compositional Knowledge) dataset}

We evaluate language models on the SICK data, which consists of 4500/4627 sentence pairs for train/test splits with the scores ranging from 1-5.The results are in the following Table, the result shows that window-mask has obvious better representation ability when using context vectors and output vectors. The proposed window-masking can be used as a language representation in terms of semantic understanding. More test results on public dataset can be benchmarked to study its performance.

\begin{table}[!htbp]
\centering
\caption{Pearson’s r × 100 results on SICK data}
\begin{tabular}{cccc} 
\toprule
\multicolumn{1}{c}{ \multirow{2}*{Method} }& \multicolumn{3}{c}{SICK-test}\\
\cline{2-4}
\multicolumn{1}{c}{}&{context}&{embed}&{output}\\

\hline
{GPT}&20.8&23.04&20.3\\

\cline{1-4}  
{window mask}&42.4&23.02&36.0\\
\cline{1-4}
{BERT}&33.5&23.5&25.9\\

\bottomrule
\end{tabular}
\end{table}

\section{Conclusion}

In this work, we propose a attention based bidirectional language model named text denoised autoencoder, in order to save the training time for bidirectional language models as well as reduce the computation time of context language representations for unsupervised applications. We conduct both reranking test and the semantic textual similarity tasks to validate the proposed method in downstream applications, the result of which demonstrate that the proposed text denoised autoencoder is apparently faster than the conventional BERT based method in terms of producing contextualized representation. Moreover, the proposed method yields context representations which have more beneficial effect for downstream applications, demonstrating its improved encoding ability when compared with that of BERT and GPT.

\bibliographystyle{unsrtnat}
\bibliography{references}  

\begin{thebibliography}{8}
\providecommand{\natexlab}[1]{#1}
\providecommand{\url}[1]{\texttt{#1}}
\expandafter\ifx\csname urlstyle\endcsname\relax
  \providecommand{\doi}[1]{doi: #1}\else
  \providecommand{\doi}{doi: \begingroup \urlstyle{rm}\Url}\fi

\bibitem[Rumelhart et~al.(1986)Rumelhart, Hinton, and
  Williams]{rumelhart1986learning}
David~E Rumelhart, Geoffrey~E Hinton, and Ronald~J Williams.
\newblock Learning representations by back-propagating errors.
\newblock \emph{nature}, 323\penalty0 (6088):\penalty0 533--536, 1986.

\bibitem[Bengio et~al.(2003)Bengio, Ducharme, Vincent, and
  Janvin]{bengio2003neural}
Yoshua Bengio, R{\'e}jean Ducharme, Pascal Vincent, and Christian Janvin.
\newblock A neural probabilistic language model.
\newblock \emph{The journal of machine learning research}, 3:\penalty0
  1137--1155, 2003.

\bibitem[Mikolov et~al.(2013)Mikolov, Sutskever, Chen, Corrado, and
  Dean]{mikolov2013distributed}
Tomas Mikolov, Ilya Sutskever, Kai Chen, Greg Corrado, and Jeffrey Dean.
\newblock Distributed representations of words and phrases and their
  compositionality.
\newblock \emph{arXiv preprint arXiv:1310.4546}, 2013.

\bibitem[Peters et~al.(2018)Peters, Neumann, Iyyer, Gardner, Clark, Lee, and
  Zettlemoyer]{peters2018deep}
Matthew~E Peters, Mark Neumann, Mohit Iyyer, Matt Gardner, Christopher Clark,
  Kenton Lee, and Luke Zettlemoyer.
\newblock Deep contextualized word representations.
\newblock \emph{arXiv preprint arXiv:1802.05365}, 2018.

\bibitem[Radford et~al.(2018)Radford, Narasimhan, Salimans, and
  Sutskever]{radford2018improving}
Alec Radford, Karthik Narasimhan, Tim Salimans, and Ilya Sutskever.
\newblock Improving language understanding by generative pre-training.
\newblock \emph{open ai}, 2018.

\bibitem[Devlin et~al.(2018)Devlin, Chang, Lee, and Toutanova]{devlin2018bert}
Jacob Devlin, Ming-Wei Chang, Kenton Lee, and Kristina Toutanova.
\newblock Bert: Pre-training of deep bidirectional transformers for language
  understanding.
\newblock \emph{arXiv preprint arXiv:1810.04805}, 2018.

\bibitem[Lan et~al.(2019)Lan, Chen, Goodman, Gimpel, Sharma, and
  Soricut]{lan2019albert}
Zhenzhong Lan, Mingda Chen, Sebastian Goodman, Kevin Gimpel, Piyush Sharma, and
  Radu Soricut.
\newblock Albert: A lite bert for self-supervised learning of language
  representations.
\newblock \emph{arXiv preprint arXiv:1909.11942}, 2019.

\bibitem[Sanh et~al.(2019)Sanh, Debut, Chaumond, and Wolf]{sanh2019distilbert}
Victor Sanh, Lysandre Debut, Julien Chaumond, and Thomas Wolf.
\newblock Distilbert, a distilled version of bert: smaller, faster, cheaper and
  lighter.
\newblock \emph{arXiv preprint arXiv:1910.01108}, 2019.

\end{thebibliography}






\end{document}